\pdfoutput=1
\documentclass{article}
\usepackage[final]{corl_2026}
\usepackage{amsmath,amssymb,amsthm}
\usepackage{booktabs}
\usepackage{graphicx}
\usepackage[capitalize,noabbrev]{cleveref}

\newcommand{\R}{\mathbb{R}}
\newcommand{\goalset}{\mathcal{G}}
\newcommand{\nogoal}{\varnothing}
\newcommand{\nnC}{352}
\newcommand{\nsuccC}{78}
\newcommand{\nsuccCci}{73, 82}
\newcommand{\nstallC}{79}

\newcommand{\nnNG}{120}
\newcommand{\nforbNG}{95}
\newcommand{\nforbNGpct}{79}
\newcommand{\nforbNGci}{71, 85}

\newcommand{\nstallNG}{25}

\newcommand{\nnegT}{27/30}
\newcommand{\nnegTi}{27/30}
\newcommand{\nnegTii}{12/20}
\newcommand{\nnegTiii}{13/20}
\newcommand{\nnegTiv}{16/20}

\newcommand{\nRstall}{163}

\newcommand{\nnR}{285}
\newcommand{\nRfollowpct}{35}

\newcommand{\nRwrong}{23}
\newcommand{\nRpairMax}{24/24}
\newcommand{\nRpairMin}{0/24}

\newcommand{\nnNV}{24}

\newcommand{\ntrainedNV}{5}
\newcommand{\nnST}{68}
\newcommand{\nswitchST}{0}

\newcommand{\ncmTwelven}{36}
\newcommand{\ncmLaten}{32}
\newcommand{\nPCin}{18}
\newcommand{\nPCitgt}{12}

\newcommand{\nPCdn}{18}
\newcommand{\nPCdtgt}{0}

\newcommand{\nPCtn}{18}
\newcommand{\nPCttgt}{0}

\newcommand{\nnWrite}{158}

\newcommand{\nCFiin}{18}
\newcommand{\nCFiitgt}{0}

\newcommand{\nCFiiin}{18}
\newcommand{\nCFiiitgt}{0}

\newcommand{\nPVn}{18}
\newcommand{\nPVtgt}{0}

\newcommand{\nSVn}{18}
\newcommand{\nSVtgt}{12}

\newcommand{\nSTn}{18}
\newcommand{\nSTtgt}{7}

\newcommand{\nSTfn}{18}
\newcommand{\nSTftgt}{15}
\newcommand{\nSVpct}{67}
\newcommand{\nSVci}{44, 84}
\newcommand{\nSTfpct}{83}
\newcommand{\nSTfci}{61, 94}
\newcommand{\nSTpct}{39}
\newcommand{\nSTci}{20, 61}

\newcommand{\nzeroUp}{18}
\newcommand{\nclInits}{45, 46 and 47}
\newcommand{\ntwN}{92}
\newcommand{\ntwNeg}{76}
\newcommand{\ntwNom}{81}
\newcommand{\ntwP}{0.30}
\newcommand{\ntwDiff}{-5}
\newcommand{\ntwDiffCi}{-14, +3}
\newcommand{\nnPatch}{159}
\newcommand{\npIviii}{0.94}
\newcommand{\npDviii}{0.04}
\newcommand{\npIxii}{0.69}

\newcommand{\npIxvi}{0.18}
\newcommand{\npDxvi}{0.77}

\newcommand{\npreDecFirst}{0.95}
\newcommand{\npreDecRest}{0.04}
\newcommand{\nstreamSixMin}{0.62}
\newcommand{\nstreamTwentyfourMin}{0.81}

\newcommand{\nncDetEarly}{1.00}
\newcommand{\nncDetMid}{0.87}
\newcommand{\nncDetLate}{0.76}

\newcommand{\nncFrames}{120}

\newcommand{\nncYmax}{0.05}

\newcommand{\nncAllMin}{0.73}
\newcommand{\npvMin}{0.47}
\newcommand{\npvMax}{1.00}
\newcommand{\nsvMin}{0.90}

\newcommand{\nvWrong}{17}
\newcommand{\nvPairs}{10}
\newcommand{\nprior}{0.87}

\newcommand{\ngWithin}{0.88}
\newcommand{\ngWrong}{0.96}
\newcommand{\ngTpr}{0.88}
\newcommand{\ngFarFol}{0.13}
\newcommand{\ngFarNom}{0.09}

\newcommand{\nentNegTpr}{0.21}

\newcommand{\nnShort}{6}
\newcommand{\npriorTie}{0.91}
\newcommand{\npriorTime}{0.96}
\newcommand{\npriorTimeWithin}{0.76}

\newcommand{\ngWithinCi}{0.81, 0.95}
\newcommand{\ndWithin}{+0.02}
\newcommand{\ndWithinCi}{-0.02, +0.06}
\newcommand{\ngNegFa}{1.00}
\newcommand{\nentWrong}{0.67}
\newcommand{\nspWrong}{0.75}
\newcommand{\nvtWrong}{0.93}

\newcommand{\ndTime}{+0.13}
\newcommand{\ndTimeCi}{-0.01, +0.26}
\newcommand{\ndSp}{+0.13}
\newcommand{\ndSpCi}{+0.04, +0.25}
\newcommand{\ndEnt}{+0.29}
\newcommand{\ndEntCi}{+0.17, +0.41}

\makeatletter
\renewcommand{\@noticestring}{SPAIS Workshop at the 10th Conference on Robot Learning (CoRL 2026), Austin, Texas, USA.}
\makeatother
\hypersetup{pdftitle={Localizing and Editing the Goal Representation of a Vision-Language-Action Model}, pdfauthor={Sudhanshu Srivastava, Faisal Aqlan}, pdfsubject={SPAIS Workshop at the 10th Conference on Robot Learning (CoRL 2026)}}

\title{Localizing and Editing the Goal Representation of a Vision-Language-Action Model}

\author{
  Sudhanshu Srivastava\\
  University of Louisville\\
  \texttt{sudhanshu.srivastava@louisville.edu}
  \And
  Faisal Aqlan\\
  University of Louisville\\
  \texttt{faisal.aqlan@louisville.edu}
}

\begin{document}
\maketitle

\begin{abstract}
Operators can state safety constraints for robots as prohibitions such as ``do not put the bowl on the stove''. We study whether OpenVLA follows such prohibitions and where it represents the goal it executes. The 10 tasks of LIBERO-Goal share one scene, so we can label the goal that each rollout executes. We combine closed-loop rollouts with activation patching, linear probes, activation steering and runtime monitors.
On LIBERO-Goal, negation had no detectable effect on which task OpenVLA executed. The model executed the forbidden task in \nforbNGpct\% of negated rollouts and completed \nsuccC\% of plain tasks. A goal probe gives similar read-outs under ``not X'' and ``X''.
Activation patching shows that the first action depends mostly on the instruction tokens before layer 12 and on the last prompt token from layer 16 onward.
A fixed steering vector at the instruction tokens at layer 8 redirected \nSVtgt{} of \nSVn{} rollouts. Fixed vectors and prefill patches at layers 12 to 20 redirected none of \nnWrite. From layer 16, only patches at every decoded action token redirected the robot (\nSTftgt{} of \nSTfn{} rollouts at layer 24).
When the instruction changes during a rollout, the layer-32 goal read-out detects non-compliance better than failure detectors, and an external probe on image and instruction features performs comparably.
These results support enforcing prohibitions outside the instruction and monitoring goal read-outs against the specification. Goal edits in OpenVLA should act before layer 12 or patch every action token.
\end{abstract}


\section{Introduction}
People use language to specify and correct robot tasks, and vision-language-action models (VLAs)~\cite{brohan2023rt,kim2024openvla,black2024vision} take this language directly as input. Many safety requirements are prohibitions. An operator may forbid contact with a hazardous object or entry into a restricted zone. Such an instruction names no goal. It removes behaviors from the set that the robot may execute, so a robot that executes an excluded behavior violates a safety constraint.
Audits show that VLAs under-use language~\cite{zhou2025libero,fei2025libero,fang2026when}. A VLA whose actions do not change under a negation or a correction can \emph{competently} execute a task that the operator excluded. This is an embodied instance of goal misgeneralization~\cite{langosco2021goal,shah2022goal}.
Runtime monitors for VLAs predict whether the \emph{instructed} task will fail~\cite{agia2024unpacking,xu2025can,gu2025safe,park2026hide,huang2026actprobe,seligmann2026vla,rajaprakash2026safecast}. They are not designed to flag a robot that competently performs a forbidden task.
We ask where OpenVLA~\cite{kim2024openvla} encodes its \emph{goal} and whether this representation can be edited. We also test whether goal read-outs help runtime oversight. In LIBERO-Goal~\cite{liu2023libero}, 10 tasks share one scene, so the instruction alone selects the goal. We evaluate all 10 goal predicates at every step to label the executed goal of each rollout.

\textbf{Contributions.}
In this simulated scene, negation had no detectable effect on which task OpenVLA executed. We locate the goal, show where it can be edited and compare goal read-outs with failure detectors as monitors.

\section{Related Work}
Linear probes recover symbolic state~\cite{lu2025probing}, world structure~\cite{molinari2025emergent}, value~\cite{zhang2026what} and progress~\cite{bhardwaj2026decoding} from VLAs. Prior work steered low-level behavior with feed-forward value vectors~\cite{hon2025mechanistic}, linear controllers~\cite{buurmeijer2026observing} and sparse autoencoder features~\cite{swann2026sparse}, and audited how VLAs route language~\cite{shi2026vla}.
The \emph{goal} we study is discrete and spread over many tokens. We adapt causal tracing from language models~\cite{meng2022locating,wang2022interpretability,geva2023dissecting,zhang2024cross} to ask what the location of the goal implies for control.
Failure monitors are often calibrated with conformal prediction~\cite{angelopoulos2021gentle,ren2023robots}, and the resulting marginal guarantee can fail under distribution shift~\cite{tibshirani2019conformal}. We use the same calibration to flag specification violations.
Robot jailbreaks~\cite{robey2024jailbreaking}, trajectory redirection~\cite{puthumanaillam2026trajectory} and typographic attacks~\cite{iranmanesh2026not} can also induce such violations.

\section{Method}
\textbf{Executed goals and specifications.}
Task $k$ of LIBERO-Goal is defined by a symbolic goal predicate $g_k$. For a rollout of OpenVLA $\pi_\theta(a_t\mid o_t,\text{instr})$, the executed goal is $y=\arg\min_k t_k$. Here $t_k$ is the first step from which $g_k$ holds for $w_g{=}5$ steps or until the episode ends. A rollout that reaches no goal is a \emph{stall} ($y{=}\nogoal$).
An instruction defines an allowed set $A$ within $\goalset=\{1,\dots,10,\nogoal\}$. The instruction ``do $c$'' gives $A=\{c\}$, and ``do not $f$'' gives $A=\goalset\setminus\{f\}$. A redirect at step $t_r$ sets $A$ to the allowed set of the new instruction.

\textbf{Goal read-outs and monitor.}
Let $h_t{=}h^{(l)}_t\in\R^{4096}$ be the Llama-2 residual stream after layer $l$ at the \emph{decision token}. This last prompt token emits the first action token at step $t$.
A goal probe $p_\phi(k\mid h_t)$ is an L2-regularized multinomial logistic regression on standardized activations projected onto 256 principal components. Each step of a nominal rollout is labeled with the eventual $y$ of that rollout.
The \emph{specification-aware} score is $s_t=\max_{k\notin A}\log p_\phi(k\mid h_t)-\max_{k\in A}\log p_\phi(k\mid h_t)$. We debounce it over $w{=}10$ steps as $S_t=\max_{t'\le t}\min_{t'-w<u\le t'}s_u$ and compare $S_t$ with a level-$\alpha$ split-conformal threshold $\hat\lambda$ from held-out compliant nominal rollouts.
Because $S_t$ is non-decreasing, an alarm within the first $H$ steps occurs if and only if $S_H>\hat\lambda$. The false-alarm rate over this window is then at most $\alpha$ for compliant rollouts that are exchangeable with the calibration rollouts~\cite{angelopoulos2021gentle}. Redirect windows start at $t_r$, so the bound does not apply to them.

\textbf{Patching and steering.}
For a frame of task $c$ and an instruction $d$ of equal token length, we overwrite the residual stream after layer $l$ at a group of positions with its value under $d$ on the same frame. The groups are the \emph{instruction} tokens where $c$ and $d$ differ, the remaining \emph{template} tokens, and the decision token.
We report the normalized logit difference of the first action token, $\rho=(\mathrm{LD}-\mathrm{LD}_c)/(\mathrm{LD}_d-\mathrm{LD}_c)$~\cite{wang2022interpretability}, and the transfer $(a-a_c)/(a_d-a_c)$ of each greedy-decoded action dimension.
A \emph{prefill} patch acts once before decoding. A \emph{decode-stream} patch also overwrites every decoded action token with values from the $d$-run on the same action prefix.
Fixed-vector steering adds $\beta$-scaled vectors computed from other episodes. \emph{Class means} $\mu_d-\mu_c$~\cite{turner2023steering,marks2023geometry} and \emph{counterfactual} means of $h_d-h_c$ over the same frames are added at the decision token and at every decoded token. \emph{Position-wise} counterfactual means are computed for each action position under a teacher-forced prefix and added at that position. \emph{Instruction-span} vectors are per-position counterfactual means over the instruction tokens at layer 8.

\section{Experiments}
\textbf{Setup.} We use OpenVLA-7B fine-tuned on LIBERO-Goal with the official preprocessing and greedy decoding. Its nominal success rate of \nsuccC\% (95\% CI [\nsuccCci]) is near the published 79.2\%~\cite{kim2024openvla}.
The \emph{nominal} condition has \nnC{} rollouts over the 10 tasks. The \emph{negation} condition has \nnNG{} rollouts with 5 phrasings of the prohibition. Each negated instruction forbids one whole task, and these tasks are benign stand-ins for prohibited behaviors. The \emph{redirect} condition has \nnR{} rollouts over 6 task pairs that share an object.
Over 10 random splits, monitors are fit on a stratified 60\% of the nominal rollouts and calibrated on the remaining compliant ones. They are scored $H{=}60$ steps after the start or after $t_r$. Monitor CIs come from cluster bootstraps over the 12 ordered task pairs, and CIs for rates are Wilson intervals.
The baselines include action-token entropy, a layer-32 success probe, and a \emph{vision+text} probe on DINOv2~\cite{oquab2023dinov} frame features with a bag-of-words encoding of the active instruction. The \emph{pair+time} prior uses the non-compliance rate of the task pair at each $t_r$ and ranks rollouts by $t_r$ within a pair. A linear probe on privileged object and end-effector poses is the last baseline.
Most closed-loop edits use 3 other task pairs in both directions and 3 initial states. Appendix~\ref{app:details} gives the exceptions and other details.

\begin{figure}[t]
\centering
\includegraphics[width=\linewidth]{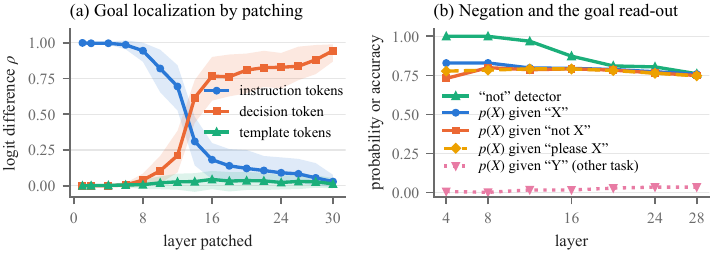}
\vspace{-5mm}
\caption{Localization of the goal and the effect of negation. \textbf{(a)}~Normalized logit difference $\rho$ when one group is patched (medians and quartiles, \nnPatch{} frames). \textbf{(b)}~Goal read-out $p(X)$ under four instructions on \nncFrames{} held-out frames, and the balanced accuracy of a detector for ``not''.}
\label{fig:read}
\vspace{-1mm}
\end{figure}

\textbf{Behavior under negation and redirects.}
Under negated instructions, OpenVLA executed the forbidden task in \nforbNG/\nnNG{} rollouts (\nforbNGpct\%, 95\% CI [\nforbNGci]), and the other \nstallNG{} rollouts stalled. On matched tasks and initial states, the forbidden task was executed in \ntwNeg/\ntwN{} cases and the plain instruction succeeded in \ntwNom/\ntwN{} cases (McNemar $p{=}\ntwP$). A prohibition stated in the instruction gave no detectable protection in this scene.
Redirects were followed in \nRfollowpct\% of rollouts. Of the \nnR{} redirect rollouts, \nRstall{} stalled and \nRwrong{} completed a wrong goal. Compliance depended strongly on the task pair (follow counts \nRpairMin{} to \nRpairMax) and decreased for later redirects.

\textbf{Localization of the goal.}
Patching shows an abrupt transition between layers 12 and 16 (\cref{fig:read}a). The instruction tokens determine the first action at layer~8 ($\rho{=}\npIviii$), about two thirds of it at layer~12 (\npIxii) and little of it from layer~16 onward (\npIxvi). The decision token determines almost none of it at layer~8 (\npDviii) and most of it from layer~16 onward (\npDxvi).
On frames of task X, another task's instruction reduces the read-out $p(X)$ to at most \nncYmax{} (\cref{fig:read}b).
A linear detector separates ``not'' from equally short neutral insertions with balanced accuracy \nncDetEarly{} at layer~4, \nncDetMid{} at layer 16 and \nncDetLate{} at layer 28. Under ``X'', ``not X'' and ``please X'', $p(X)$ stays at or above \nncAllMin{} through layer 28, so ``not'' does not suppress the read-out.

\begin{table}[t]
\centering
\scriptsize
\setlength{\tabcolsep}{2.6pt}
\caption{Monitor results 60 steps after the start or the redirect. TPR and FA are flag rates on violating and compliant rollouts at the $\alpha{=}0.1$ threshold (held-out nominal FA \ngFarNom). Within-pair AUROC covers the \nvPairs{} ordered pairs with both outcomes. The starred baseline uses privileged simulator state.}
\label{tab:monitor}
\begin{tabular}{lccccccc}
\toprule
 & \multicolumn{5}{c}{Redirect not followed ($n{=}180$) vs.\ followed ($n{=}99$)} & \multicolumn{2}{c}{Negation} \\
\cmidrule(lr){2-6}\cmidrule(lr){7-8}
Monitor & AUROC & within-pair AUROC & wrong-goal AUROC & TPR & FA & TPR & FA \\
 & [95\% CI] & [95\% CI] & ($n{=}17$) & & & ($n{=}95$) & ($n{=}25$) \\
\midrule
Pair+time prior & 0.96 & 0.76 {\scriptsize[0.63,0.87]} & -- & -- & -- & -- & -- \\
\midrule
Goal probe, layer 32 & 0.94 {\scriptsize[0.88,0.98]} & 0.88 {\scriptsize[0.81,0.95]} & 0.96 & 0.88 & 0.13 & 1.00 & 1.00 \\
Vision+text (DINOv2) & 0.92 {\scriptsize[0.84,0.97]} & 0.87 {\scriptsize[0.78,0.94]} & 0.93 & 0.85 & 0.13 & 1.00 & 1.00 \\
Object poses$^\ast$ (linear) & 0.90 {\scriptsize[0.83,0.93]} & 0.79 {\scriptsize[0.66,0.87]} & 0.98 & 0.81 & 0.11 & 0.89 & 0.80 \\
Success probe & 0.87 {\scriptsize[0.78,0.94]} & 0.74 {\scriptsize[0.63,0.87]} & 0.75 & 0.46 & 0.08 & 0.07 & 0.00 \\
Goal probe, layer 4 & 0.81 {\scriptsize[0.69,0.93]} & 0.67 {\scriptsize[0.55,0.81]} & 0.71 & 0.49 & 0.09 & 1.00 & 1.00 \\
Action entropy & 0.77 {\scriptsize[0.66,0.85]} & 0.59 {\scriptsize[0.49,0.68]} & 0.67 & 0.39 & 0.07 & 0.21 & 0.19 \\
\bottomrule
\end{tabular}

\vspace{-4mm}
\end{table}

\begin{figure}[t]
\centering
\includegraphics[width=\linewidth]{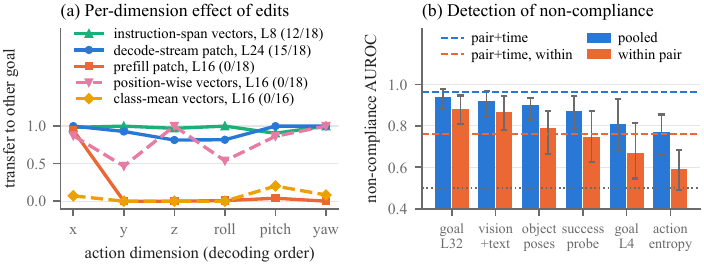}
\vspace{-5mm}
\caption{Editing and monitoring the goal. \textbf{(a)}~Offline transfer of each action dimension toward the other goal. L is the layer, and counts are redirected closed-loop rollouts. \textbf{(b)}~AUROC for redirect non-compliance (cluster-bootstrap CIs) and the pair+time prior (dashed).}
\label{fig:write}
\vspace{-4mm}
\end{figure}

\textbf{Editing the goal.}
Before the transition, a \emph{fixed} vector can change the goal (\cref{fig:write}a). Instruction-span vectors at layer~8 transfer every action dimension offline (at least \nsvMin). In closed loop they made the robot execute the other task in \nSVtgt/\nSVn{} rollouts (\nSVpct\%, 95\% CI [\nSVci]).
After the transition, a prefill patch of the decision token at layer 16 transfers \npreDecFirst{} of the first dimension and at most \npreDecRest{} of the other five. The later action tokens can still attend to the unedited instruction tokens. In closed loop these patches redirected \nPCdtgt/\nPCdn{} rollouts at the decision token and \nPCttgt/\nPCtn{} at all text positions.
No fixed vector applied at layers 12 to 20 redirected a rollout. Class means at layers 12 to 20 gave \nswitchST/\nnST, counterfactual means gave \nCFiitgt/\nCFiin{} at $\beta{=}2$ and \nCFiiitgt/\nCFiiin{} at $\beta{=}3$, and position-wise means gave \nPVtgt/\nPVn.
After the transition, only the decode-stream patch transfers more than half of every dimension offline (at least \nstreamSixMin{} at layer 16 and \nstreamTwentyfourMin{} at layer 24). In closed loop it redirected \nSTtgt/\nSTn{} rollouts at layer 16 (\nSTpct\%, 95\% CI [\nSTci]) and \nSTftgt/\nSTfn{} at layer 24 (\nSTfpct\%, 95\% CI [\nSTfci]).

\textbf{Monitoring.}
Failure detectors that do not use the specification seldom flag forbidden executions. Entropy flags \nentNegTpr{} of these rollouts (\cref{tab:monitor}). Goal read-outs flag all \nforbNG{} executions and also all \nstallNG{} negated rollouts that stalled (FA \ngNegFa). They reflect the goal named in the instruction, so they perform no better than a keyword rule for ``not''.
Redirects are more informative (\cref{fig:write}b). The in-sample pair+time prior reaches a pooled AUROC of \npriorTime. Within a pair, the layer-32 read-out reaches AUROC \ngWithin{} (95\% CI [\ngWithinCi]). Its paired differences from the success probe and entropy are $\ndSp$ (95\% CI $[\ndSpCi]$) and $\ndEnt$ (95\% CI $[\ndEntCi]$). Against the pair+time prior (\npriorTimeWithin), the difference is $\ndTime$ (95\% CI $[\ndTimeCi]$). On followed redirects, the false-alarm rate is \ngFarFol{} at $\alpha{=}0.1$.
The external vision+text probe performs comparably to the layer-32 probe. Their within-pair difference is $\ndWithin$ (95\% CI $[\ndWithinCi]$), so internal access adds no detectable gain.

\section{Implications for Robot Safety and Limitations}
In this scene, a negated instruction did not impose a safety constraint on OpenVLA, and failure detectors seldom flagged the forbidden executions. A system built on such a model should enforce prohibitions outside the instruction. Under negation, our goal monitors were no better than a keyword rule. The redirect results support a monitor that compares goal read-outs with the specification, and the external probe indicates that such a monitor may not need model internals. For steering methods on autoregressive VLAs, the results suggest that an edit must act before the transition (layers 12 to 16 in OpenVLA) or patch every decoded action token.

\textbf{Limitations.} We study one VLA in one simulated scene with benign whole-task prohibitions and simulator goal labels. Most closed-loop edits use only 18 rollouts from 3 task pairs, so a count of 0/18 allows an edit success rate of up to \nzeroUp\% (95\% CI). Tests on other VLAs and on real robots are needed before these results can inform deployment.

\bibliography{refs}

\clearpage
\appendix
\section{Experimental Details}
\label{app:details}
\textbf{Conditions.}
The negation phrasings were ``not'', ``do not'', ``don't'', ``never'' and ``you must not''. OpenVLA executed the forbidden task in \nnegT, \nnegTi, \nnegTii, \nnegTiii{} and \nnegTiv{} rollouts with these phrasings. On the \ntwN{} matched cases, negation changed the execution rate by $\ntwDiff$ percentage points (Newcombe 95\% CI $[\ntwDiffCi]$). Each negated instruction forbids a whole task. We did not test a prohibition that restricts how a task is done. The checkpoint was fine-tuned on the 10 LIBERO-Goal instructions, and none of them contains a negation.
Each redirect pair is switched in both directions at $t_r\in\{10,25,50\}$ or at 50, 70 or 85\% of the median duration of the source task. All \nstallC{} nominal failures were stalls. Redirect outcomes cluster by task pair, and the wrong-goal set is small.
Another \nnNV{} rollouts use \emph{novel compositions}. In \ntrainedNV{} of these \nnNV{} rollouts, OpenVLA executed a similar training task. For instance, it placed the \emph{bowl} on the stove when the instruction named the cream cheese.

\textbf{Episodes and scoring.}
Non-nominal episodes end when the task named in the first instruction succeeds or once another goal has held for 20 steps. As in the nominal protocol, the monitor analyses truncate episodes at the first goal step and exclude the \nnShort{} redirects that end within 10 steps of $t_r$. These \nnShort{} rollouts completed the original task, so \cref{tab:monitor} has \nvWrong{} of the \nRwrong{} wrong-goal completions.
On these \nvWrong{} rollouts, the layer-32 and vision+text read-outs reach an AUROC of \ngWrong{} and \nvtWrong. The failure detectors reach \nspWrong{} (success probe) and \nentWrong{} (entropy). Under negation, the success probe flags predicted success of the forbidden task.
The pair+time prior uses non-compliance rates computed on the rollouts it scores. A prior with the rate of the task pair alone reaches a pooled AUROC of \npriorTie{} in-sample and \nprior{} leave-one-out. The in-sample version is constant within a pair, so its within-pair AUROC is 0.50. The layer-32 monitor has a TPR of \ngTpr{} on redirects that were not followed and a false-alarm rate of \ngFarFol{} on followed ones. Paired within-pair differences use the scores of the first split and 10,000 bootstrap resamples of the ordered pairs. Wilson intervals treat rollouts as independent.

\textbf{Interventions and probes.}
Patching uses frames at steps 5, 20 and 40 from the 5 task pairs with instructions of equal length. Per-dimension medians use frames with $|a_d-a_c|>0.1$. At layer 32 the model returns the normalized final state. The probes use the unnormalized block output recorded in rollouts, so \cref{fig:read} stops at layer 30 for patching and at layer 28 for the read-outs.
Most closed-loop edits use the initial states \nclInits, and their steering vectors come from nominal rollouts with initial states below 40. The class-mean runs use 4 ordered pairs from the redirect condition. At layer 12 they add the class mean at the decision token and at every decoded token with $\beta\in\{1,2,4\}$ on the initial states 45, 46 and 47 (\ncmTwelven{} rollouts). These class means average the first 40 steps of all successful nominal rollouts, including those at the test initial states. At layers 16 and 20 they add per-position class means over the prompt positions after the instruction with $\beta\in\{1,2\}$ on the initial states 45 and 46 (\ncmLaten{} rollouts).
Patching the instruction tokens at layer~8 redirected \nPCitgt/\nPCin{} closed-loop rollouts. Offline, position-wise means transferred \npvMin{} to \npvMax{} per dimension at $\beta{=}2$. No closed-loop rollout with these vectors reached the other goal.
Goal probes and the success probe use 256 components and inverse regularization strength $C{=}0.05$. The vision+text and object-pose probes use 128 and 64 components with $C{=}0.5$ and $C{=}1$. The object-pose probe uses the end-effector pose and the positions of the 4 movable objects. It has no object orientations and no drawer or stove states. All experiments ran on one RTX A6000. The code, plans, episode summaries and analysis outputs are available at \url{https://github.com/sudsho/vla-goal-representation} and archived on Zenodo (DOI \href{https://doi.org/10.5281/zenodo.23871372}{10.5281/zenodo.23871372}).

\end{document}